\pdfoutput=1

\documentclass[11pt]{article}

\usepackage[]{naacl2021}

\usepackage{times}
\usepackage{latexsym}

\usepackage[T1]{fontenc}
\usepackage[utf8]{inputenc}

\usepackage{microtype}

\usepackage{graphicx}
\usepackage{amsmath}
\usepackage{amsthm}
\usepackage{booktabs}
\usepackage{algorithm}
\usepackage{algorithmic}
\usepackage{xspace}
\usepackage{CJKutf8}
\usepackage{microtype}
\usepackage{subcaption}
\usepackage{blindtext}
\usepackage{todonotes}

\usepackage{amssymb}%
\usepackage{pifont}%
\newcommand{\cmark}{\ding{51}}%
\newcommand{\xmark}{\ding{55}}%

\usepackage{multirow}
\usepackage{tabularx}

\newcommand{\ie}{i.e.,\xspace}
\newcommand{\eg}{e.g.,\xspace}
\newcommand{\eat}[1]{}

\newcommand{\baby}{\texttt{DogWhistle}\xspace}
\newcommand{\zh}[1]{\begin{CJK}{UTF8}{gbsn}#1\end{CJK}}

\title{Blow the Dog Whistle: A Chinese Dataset for Cant Understanding \\ with Common Sense and World Knowledge }

\author{Canwen Xu$^1$\thanks{\ \ Equal Contribution. Work done at Microsoft Research Asia.} , Wangchunshu Zhou$^{2*}$, Tao Ge$^3$, Ke Xu$^2$, Julian McAuley$^1$, Furu Wei$^3$\\
 $^1$ University of California, San Diego
 $^2$ Beihang University
 $^3$ Microsoft Research Asia \\
 $^1$\texttt{\{cxu,jmcauley\}@ucsd.edu}  $^3$\texttt{\{tage,fuwei\}@microsoft.com} \\
 $^2$\texttt{zhouwangchunshu@buaa.edu.cn,kexu@nlsde.buaa.edu.cn} \\
}

\date{}

\begin{document}
\maketitle
\begin{abstract}
\emph{Cant} is important for understanding advertising, comedies and dog-whistle politics. However, computational research on cant is hindered by a lack of available datasets. In this paper, we propose a large and diverse Chinese
dataset for creating and understanding cant from a computational linguistics perspective.
We formulate a task for cant understanding and provide both quantitative and qualitative analysis for tested word embedding similarity and pretrained language models. 
Experiments suggest that 
such a task
requires deep language understanding, common sense,
and
world knowledge
and
thus can be
a good
testbed for pretrained language models and help models perform better on other tasks.\footnote{The code is available at \url{https://github.com/JetRunner/dogwhistle}. The data and leaderboard are available at \url{https://competitions.codalab.org/competitions/30451}. }
\end{abstract}

\section{Introduction}
A cant\footnote{\url{https://en.wikipedia.org/wiki/Cant_(language)}} (also known as doublespeak, cryptolect, argot, anti-language or secret language) is the jargon or language of a group, often employed to exclude or mislead people outside the group~\cite{mcarthur2018oxford}. Cant is crucial for understanding advertising~\cite{dieterich1974public} and both ancient and modern comedy~\cite{sommerstein1999anatomy,prasetyo2019euphemism}. Also, it is the cornerstone for infamous dog-whistle politics~\cite{lopez2015dog,albertson2015dog}. 
Here, we summarize the key elements for cant: (1) Both a cant and its reference (\ie \textit{hidden word}) should be in the form of common natural text (not another symbol system, \eg Morse code). (2) There is some shared information between the cant users (\ie \textit{the insiders}) that is not provided to the people outside the group. (3) A cant should be deceptive and remain undetected to avoid being decrypted by people outside the group (\ie \textit{the outsiders}). These elements make the creation and understanding of cant subtle and hard to observe~\cite{taylor1974terms}. To the best of our knowledge, currently there are very few resources available for the research of cant.

\begin{figure*}[t]
     \centering
     \begin{subfigure}[t]{0.49\textwidth}
         \centering
         \includegraphics[width=\textwidth]{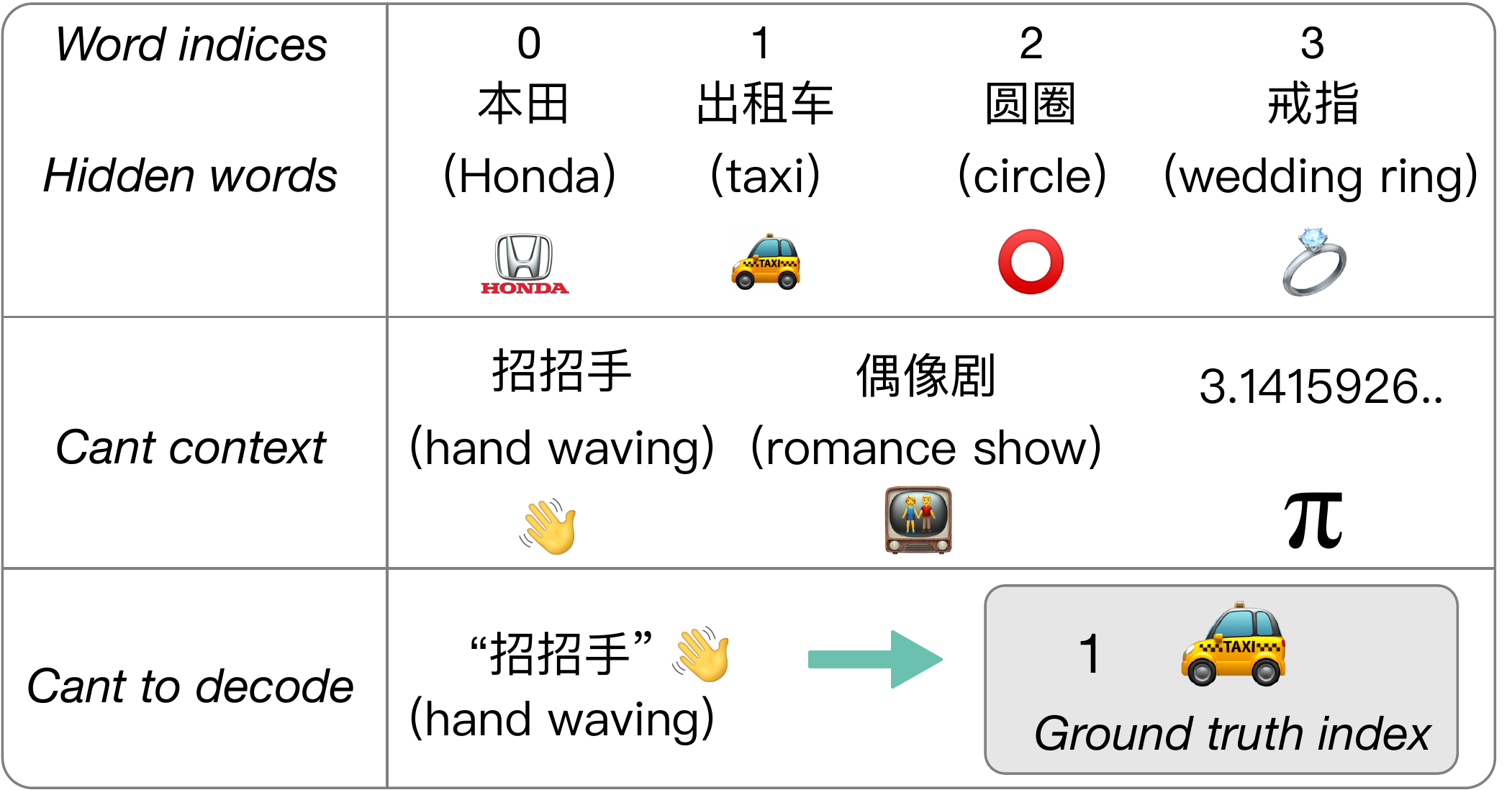}
         \caption{\textit{Insider} subtask. In this subtask, we mimic communication between insiders. The input (white background) is hidden words, cant context and a cant to decode. The model should output the index of the predicted hidden word (gray background). The hidden words are visible in this subtask.}
         \label{fig:teammate}
     \end{subfigure}
     \hfill
     \begin{subfigure}[t]{0.49\textwidth}
         \centering
         \includegraphics[width=\textwidth]{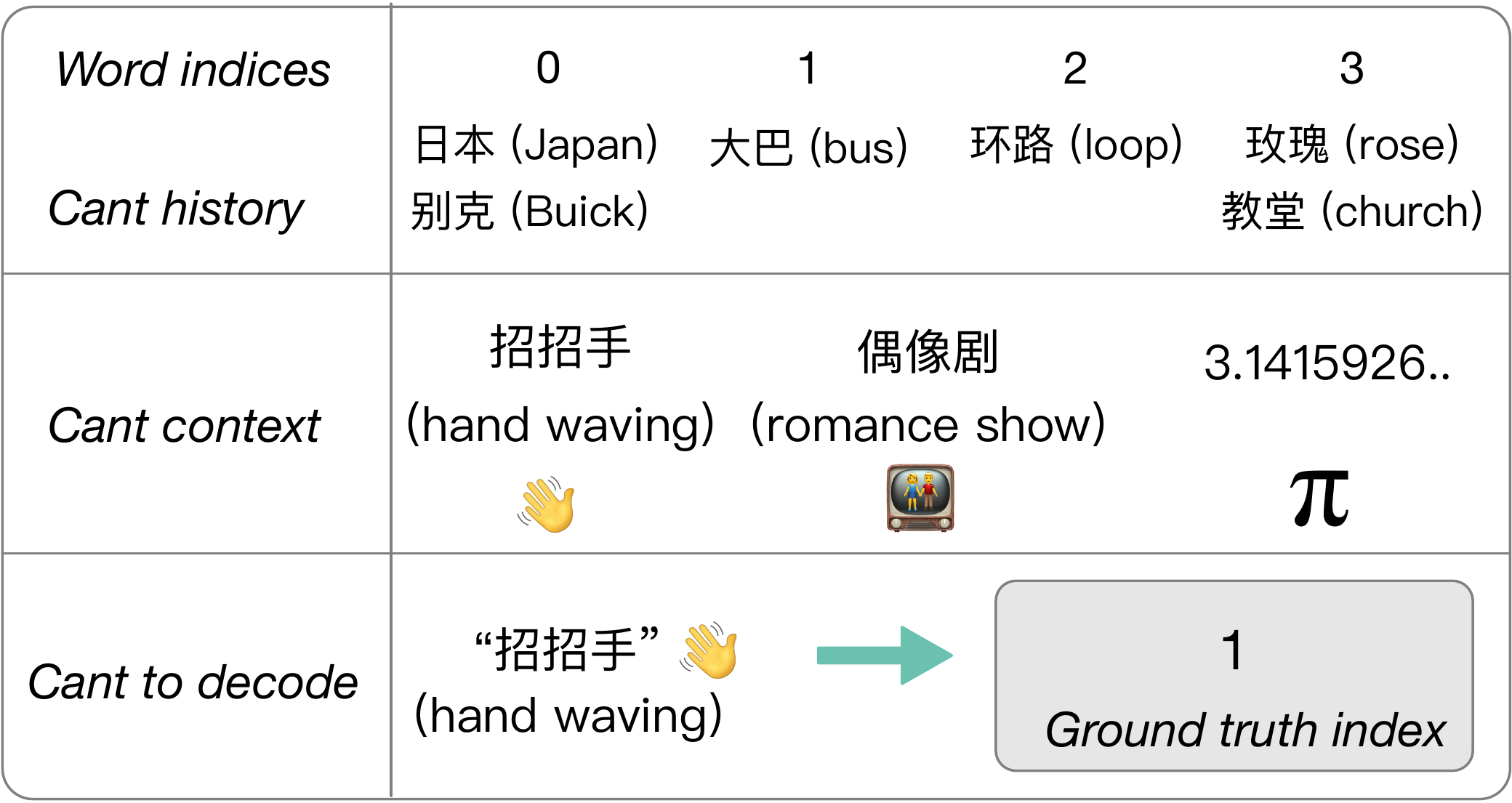}
         \caption{\textit{Outsider} subtask. In this subtask, an outsider tries to decrypt the communication by reading the cant history from previous rounds. The input is cant histories, cant context and a cant to decode (white background). The model should output the index of the predicted cant history (gray background). The hidden words are not visible in this subtask.}
         \label{fig:adversary}
     \end{subfigure}
        \caption{Input and output examples of the two subtasks of \baby. See Appendix A for more examples.}
        \label{fig:example}
\end{figure*}

In this paper, 
we create a dataset for studying cant, \baby,
centered around the aforementioned key elements (examples shown in Figure \ref{fig:example}). We collect the data with a well-designed online game under a player-versus-player setting (see Section \ref{sec:game}).
The dataset includes abundant and diverse cant for a wide spectrum of hidden words.
We find that cant understanding requires a deep understanding of language, common sense and world knowledge, making it a good testbed for next-generation pretrained language models. Our dataset also serves as a timely and complex language resource that can help models perform better on other tasks through Intermediate Task Transfer~\cite{intermediate}.

\section{Related Work}
The use of cant has long been studied in linguistics research~\cite{pei1973double,pulley1994doublespeak,albertson2006dog,squires2010enregistering,howdogwhistleswork,henderson2019dogwhistles,henderson2019dogwhistles2,bhat2020covert}. However, due to a lack of language resources, there are few studies in computational linguistics research. \citet{henderson-mccready-2020-towards} attempted to model the dog-whistle communications with a functional, agent-based method. 

As a related topic in computational linguistics, some previous studies investigate coded names in human language. 
\citet{zhang2014be} analyzed and generated coded names of public figures. \citet{zhang2015context} designed an automatic system to decode the coded names. \citet{huang2017knowledge} exploited a knowledge graph to identify coded names. \citet{huang2019multimodel} leveraged multi-modal information to align coded names with their references. 

Our work differs from 
the above
in the following ways: (1) Previous studies focused on coded names for public figures; the source and variety of these coded names is limited. 
The hidden words in our dataset are sampled from a common dictionary and are of high diversity. 
(2) The coded names in previous studies are used by users to bypass a censor
(mostly a rule-based automatic text matching system). Conversely, our data are collected under an adversarial setting, pressuring users to mislead human adversaries. Thus, our work is ideal for evaluating recent progress on Natural Language Understanding (NLU)~\cite{bert,albert,roberta,ernie,bot,pabee,unihanlm}.

\section{Data Collection}
\begin{figure}[t]
         \centering
         \includegraphics[width=\columnwidth]{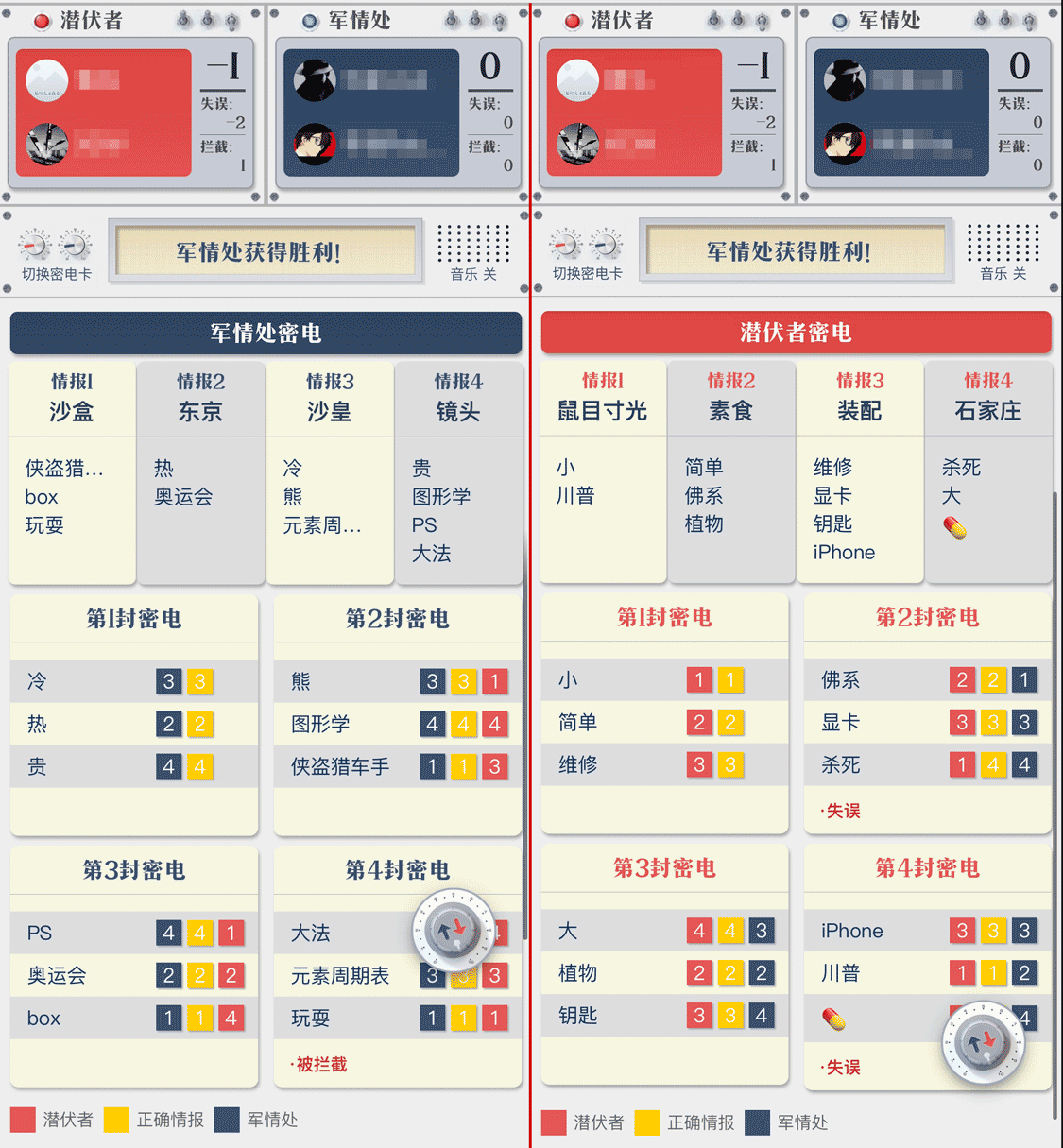}
         \caption{Screenshot of the user interface. The left and right halves of the screenshot are the screens for the two teams, respectively. The top section is the teams' scores. The middle section contains the hidden words and cant history. The bottom section is the cant to decode for each round. }
        \label{fig:screenshot}
\end{figure}

Previous studies~\cite{game1,game2} reveal that gamification can often improve the quality of collected data.
Instead of collecting data from the wild like most datasets~\cite{zhang2014be,zhang2015context,matinf}, we
collect 
the data from 
historical
game records of \textit{Decrypto Online}, a well-designed online board game. The screenshot of the user interface is shown in Figure \ref{fig:screenshot}.

\subsection{Game Design}
\label{sec:game}
The game design is adapted from the board game \textit{Decrypto}.\footnote{We recommend this video showing how to play the game: \url{https://youtu.be/2DBg7Z2-pQ4}}
Four players (\eg A, B, C and D) are divided into two teams
(\eg A and B vs.~C and D), with each trying to correctly interpret the cant presented to them by their teammates while cracking the codes they intercept from the opposing team.

\zh{
In more detail, each team has their own screen, and in this screen there are four words numbered 0-3. Both players on the same team can see their own words while hiding the words from the opposing team. In the first round, each team does the following: One team member receives a randomly generated message that shows three of the digits 0-3 in some order, \eg 3-1-0. They then give cant that their teammates must use to guess this message. For example, if A and B's four words are ``本田'' (Honda), ``出租车'' (taxi), ``圆圈'' (circle), and ``戒指'' (wedding ring), then A might say ``招招手-偶像剧-3.14" 
(``hand waving''-``romance show''-3.14) and hope that their teammate B can correctly map those cant to 0-2-1. If B guesses incorrectly, the team would receive one ``failure mark''.

Starting in the second round, a member of each team must again give a clue about their words to match a given three-digit message. One member from the other team (\eg C) then attempts to guess the message. Taking Figure \ref{fig:adversary} as an example, based on the cant histories from previous rounds, C can roughly guess the code is 0-2-1. If C is correct, C and D would receive one ``success mark''. After every round, the real messages that both teams were trying to pass will be revealed.
}

The rounds continue until a team collects either its second success mark (to win the game) or its second failure mark (to lose the game).

\subsection{Additional Rules and Restrictions}
The participants are explicitly asked not to create a cant based on its position, length, and abbreviation. That is to say, to mimic the creation of cant, we emphasize the importance of semantics instead of the morphology. 
To enforce this, all input that contains the same character as in one of the four words will be automatically rejected. As emojis have been playing an important role in online communications nowadays~\cite{chen2019emoji}, emojis are allowed as valid input.

\begin{table}[t]
\small
\centering
  \begin{tabular}{l|rrr}
  \toprule
    & train & dev & test\\
    \midrule
    \# games & 9,817 & 1,161 & 1,143 \\
    \# rounds & 76,740& 9,593 &9,592 \\
    \# word comb. & 18,832 & 2,243 & 2,220 \\ 
    \# uniq. words & 1,878 & 1,809 & 1,820\\
    \# cant & 230,220 & 28,779 & 28,776\\
    \midrule
    avg. word len. & 2.11 & 2.12 & 2.13 \\
    avg. cant len. & 2.10 & 2.10 & 2.09 \\
    \bottomrule
  \end{tabular}
  \caption{\label{stats}Statistics of our collected \baby dataset. }
\end{table}

\subsection{Data Cleaning and Split}
For data cleaning, we remove all rounds with an empty cant. We also exclude rounds where the player fails to write a cant within the given time limit (one minute). We randomly split the data into training, development and test sets with an 8:1:1 ratio, 
such that all rounds
of a game are in the same split. 
We also ensure there is no overlapping combination of hidden words between splits. We show the statistics of the training, development and test sets in Table~\ref{stats}.
In contrast to 288k cant phrases for 1.9k hidden words in our dataset, data collected by previous studies~\cite{zhang2014be,zhang2015context,huang2017knowledge} are quite small, often containing hundreds of coded names for a small set of entities.

\section{Experiments and Analysis}
\subsection{Task Formulation}
As shown in Figure~\ref{fig:example}, we have subtasks named \textit{insider} and \textit{outsider}, respectively. For the \textit{insider} subtask, we try to decode the cant to one of the hidden words. For the \textit{outsider} subtask, the hidden words are invisible and the goal is to decrypt the messages based on the communication history.
We formulate the task of decoding the cant in a similar format to multi-choice reading comprehension tasks~\cite{race,swag,arc}. We consider the cant context and the cant to decode as the ``context'' and ``question'' (respectively) as in multi-choice reading comprehension tasks. For the candidate answers, we use the hidden words and the set of cant histories for the insider subtask and the outsider subtask, respectively.
\subsection{Baselines}
\paragraph{Word Embedding Similarity}
Our task is naturally similar to the task of word similarity~\cite{semeval124}.
We select FastText~\cite{grave2018learning}, SGNS~\cite{analogical} (trained with mixed large corpus), DSG~\cite{tencentemb} and VCWE~\cite{vcwe} as word embedding baselines.
For each word embedding baseline, we first check if the cant is in the vocabulary; if it is not, we try to use a word tokenizer\footnote{We use Jieba, a popular Chinese tokenizer: \url{https://github.com/fxsjy/jieba}} to break it into words. If there is still any out-of-vocabulary token, we then break it into characters. For the \textit{insider} subtask, we take the average of the word vectors to represent the cant and select the hidden word with the smallest cosine distance in the embedding space. For the \textit{outsider} subtask, we take the average of the history cant for each hidden word as the representation. Then we predict the label by selecting the smallest distance between the representation of the cant and the history cant. Note that for word embedding baselines, the cant context is omitted and the evaluation is under a zero-shot setting (without any training).

\paragraph{Pretrained Language Models} 
We use BERT~\cite{bert}, RoBERTa~\cite{roberta}, ALBERT~\cite{albert}, and Baidu ERNIE~\cite{ernie} as baselines.\footnote{The pretrained weights for BERT are from the official BERT repository: \url{https://github.com/google-research/bert}. Pretrained weights for other models are provided by CLUE: \url{https://github.com/CLUEbenchmark/CLUEPretrainedModels}.} The implementation is based on Hugging Face's Transformers~\cite{huggingface}.
Specifically, for the \textit{insider} subtask, we construct the input sequence for each choice by concatenating its context, cant, and candidate hidden words with a special token \texttt{[SEP]}. We then concatenate the input sequences for all candidate hidden words with \texttt{[SEP]} and feed it into a BERT-like model. Finally, we use the hidden representation of the first token \texttt{[CLS]} to output the final prediction with a linear layer. For the \textit{outsider} subtask, we replace the hidden words with the cant history. We fine-tune the models on the training set and report the results on the development and test sets. We use Adam~\cite{adam} with a learning rate searched over \{2e-5, 3e-5, 5e-5\} and a batch size of 64 to fine-tune the models for 3 epochs. We warm-up the learning rate for the first 10\% of steps.

\begin{table}[t]
\centering
  \resizebox{1.\linewidth}{!}{\begin{tabular}{l|cc|cc}
  \toprule
   \multirow{2}{*}{Model} & \multicolumn{2}{c|}{\textit{Insider}} & \multicolumn{2}{c}{\textit{Outsider}}\\
   & dev & test & dev & test \\
    \midrule
    Human Performance & 87.5 & 88.9 & 43.1 & 43.1 \\
    Random Guessing & 25.0 & 25.0 & 25.0 & 25.0 \\
    \midrule
    FastText (300D)~\shortcite{grave2018learning} & 52.6 & 53.3 & 29.8 & 30.3 \\
    SGNS (300D,large)~\shortcite{analogical} & 52.3 & 52.3 & 30.6 & 30.8 \\
    DSG (200D)~\shortcite{tencentemb} & 56.3 & 56.2 & 31.4 & 31.4 \\
    VCWE (50D)~\shortcite{vcwe} & 46.0 & 46.2 & 28.0 & 28.0 \\
    \midrule
    BERT-base~\shortcite{bert} & 73.5 & 74.1 & 33.7 & 33.7 \\
    RoBERTa-base~\shortcite{roberta} & 73.5 & 74.1 & 34.0 & 34.1 \\
    ALBERT-base~\shortcite{albert} & 72.6 & 73.0 & 33.6 & 33.7 \\
    ERNIE-base~\shortcite{ernie} & 73.4 & 73.9 & 34.0 & 34.1 \\
    RoBERTa-large~\shortcite{roberta} & 74.8 & 75.4 & 34.2 & 34.3 \\
    ALBERT-xxlarge~\shortcite{albert} & 75.4 & 76.1 & 34.6 & 34.6 \\
    \bottomrule
  \end{tabular}}
  \caption{\label{tab:res}Accuracy scores of human performance and baselines for the two subtasks of \baby, \textit{insider} and \textit{outsider}. For word embedding baselines, the number of dimensions is marked, \eg (300D). }
\end{table}

\zh{
\begin{table*}[t]
\centering
  \resizebox{\textwidth}{!}{
  \begin{tabular}{c|c|c|c|cc}
  \toprule
  & \textbf{Hidden words} & \textbf{Cant context} & \textbf{Cant to decode} & \textbf{BERT} & \textbf{Human} \\
  \midrule
   \multirow{2}{*}{\#1} & 合作, 死神, 密码, 机械 & 黑人抬棺, 007, 握手 & 黑人抬棺 & 密码 \xmark & 死神 \cmark \\ 
& cooperation, Grim Reaper, password, machinery & Dancing Pallbearers, 007, handshaking & Dancing Pallbearers & password & Grim Reaper \\
\midrule
   \multirow{2}{*}{\#2} & 合作, 死神, 密码, 机械 & 黑人抬棺, 007, 握手 & \multirow{2}{*}{007} & 死神 \xmark & 密码 \cmark \\ 
& cooperation, Grim Reaper, password, machinery & Dancing Pallbearers, 007, handshaking & & Grim Reaper & password\\
\midrule
\multirow{2}{*}{\#3} & 破产, 日历, 轴, 熊孩子 & 酱油, 零, 字 & 酱油 & 日历 \xmark & 熊孩子 \cmark \\
& bankruptcy, calendar, kids & sauce, zero, digits & sauce & calendar & kids \\

    \bottomrule
  \end{tabular}
  }
  \caption{\label{tab:casestudy} Some cases that BERT fails to predict but that human players predict correctly for the \textit{insider} subtask.}
\end{table*}
}

\subsection{Quantitative Analysis}
We show the experimental results in Table \ref{tab:res}. For word embedding similarity baselines, DSG~\cite{tencentemb}, which is trained with mixed characters, words and n-grams on a diverse large corpus, drastically outperforms other word embeddings. For pretrained language models, large-size models, with more computational capacity, remarkably outperform base-size models on the \textit{insider} subtask. Both RoBERTa-base and ERNIE-base outperform BERT-base while ALBERT-base, which employs parameter sharing, slightly underperforms BERT on both tasks. Notably, the best-performing model still trails human performance by a large margin of $12.8$ and $8.5$ on the \textit{insider} and \textit{outsider} subtasks, respectively. It indicates that \baby is a very challenging dataset, providing a new battleground for next-generation pretrained language models.

\subsection{Qualitative Analysis}
We list some representative samples that BERT fails to predict but that are correctly predicted by human players in Table~\ref{tab:casestudy}. For example \#1, ``Dancing Pallbearers''\footnote{\url{https://en.wikipedia.org/wiki/Dancing_Pallbearers}} is a recent meme that went viral after the release of the models. Thus, it is 
likely
that the pretrained models have 
little
knowledge about 
the subject.
For example \#2, ``007'' refers to James Bond films\footnote{\url{https://en.wikipedia.org/wiki/James_Bond}}, in which the protagonist often cracks passwords in a mission. This kind of reasoning requires a high understanding of world knowledge instead of overfitting shallow lexical features, which 
has been
pointed out as a major drawback in natural language inference~\cite{sem18adam,acl2019selection}. For example \#3, \zh{``孩子都可以打酱油了''} (the 
child
can buy sauce) is a Chinese slang that means a 
child
has grown up. To successfully predict this example, the model must have extensive knowledge 
of
the language.

\begin{table}[t]
\centering
  \resizebox{1.\linewidth}{!}{
  \begin{tabular}{l|cc|cc}
  \toprule
   \multirow{2}{*}{Model} & \multicolumn{2}{c|}{AFQMC} & \multicolumn{2}{c}{LCQMC} \\
   & orig. & trans. & orig. & trans. \\
    \midrule
    BERT-base~\shortcite{bert} & 74.2 & \textbf{74.5} (+0.3)& 89.4 & \textbf{89.7} (+0.3)  \\
    RoBERTa-base~\shortcite{roberta} & 73.8 & \textbf{74.4} (+0.6)  & 89.2 & \textbf{89.7} (+0.5) \\
    RoBERTa-large~\shortcite{roberta} & 74.3 & \textbf{74.8} (+0.5) & 89.8 & \textbf{90.0} (+0.2) \\
    \bottomrule
  \end{tabular}
  }
  \caption{\label{tab:intermediate}Accuracy scores \textit{(dev set)} of the original performance and intermediate-task transfer performance.}
\end{table}

\subsection{Intermediate-Task Transfer}
Intermediate-Task Transfer Learning~\cite{intermediate} exploits an \textit{intermediate} task to improve the performance of a model on the target task. As we analyzed before, \baby contains rich world knowledge and requires high-level reasoning.  Therefore, we can strengthen the ability of a model by leveraging our dataset as an intermediate task. Specifically, we transfer \baby for a semantic similarity task. We first fine-tune the models on the \textit{insider} subtask, then re-finetune the models on two real-world semantic matching datasets, Ant Financial Question Matching Corpus (AFQMC)~\cite{clue} and Large-scale Chinese Question Matching Corpus (LCQMC)~\cite{lcqmc}.
As shown in Table~\ref{tab:intermediate}, on both datasets, \baby helps models significantly obtain better performance ($p<0.05$).

\section{Conclusion and Future Work}
In this paper, we propose \baby, a new Chinese dataset for cant creation, understanding and decryption. We evaluate word embeddings and pretrained language models on the dataset. The gap between human performance and 
model results
indicates that our dataset is challenging and 
promising
for evaluating new pretrained language models. For future work, we 
plan to leverage this dataset to
train agents to compete against each other,
to better understand verbal intelligence and teach agents to reason, guess and deceive in the form of natural language to make new progress at higher levels of World Scope~\cite{bisk2020experience}.
\section*{Ethical Considerations}
During data collection, the game has a guideline that asks the players not to use any offensive content when playing the game. However, like all user-generated language resources, there would inevitably be bias and stereotyping in the dataset. We consider this as a double-edged sword, which provides opportunities for computational social science research of bias in human language, but also requires responsible use of these data. We would also like to warn that there would inevitably be potentially toxic or offensive contents in the dataset.
Likewise, this dataset could be abused to generate dog-whistle phrases and political propaganda;
Being aware of the risks, we have set terms to restrict the use to be for research purposes only.

\section*{Acknowledgments}
We would like to sincerely thank Ren Wuming, the game developer of Decrypto Online, for his full support for this research. We appreciate all anonymous reviewers, especially the meta-reviewer, for their insightful comments. Canwen wants to thank all members of the Board Game Club of Microsoft Research Asia for the inspiration. Tao Ge is the corresponding author.
\bibliography{custom}

\begin{thebibliography}{47}
\expandafter\ifx\csname natexlab\endcsname\relax\def\natexlab#1{#1}\fi

\bibitem[{Albertson(2006)}]{albertson2006dog}
Bethany Albertson. 2006.
\newblock Dog whistle politics, coded communication, and religious appeals.
\newblock \emph{American Political Science Association and International
  Society of Political Psychology. Princeton, NJ: Princeton University}.

\bibitem[{Albertson(2015)}]{albertson2015dog}
Bethany~L Albertson. 2015.
\newblock Dog-whistle politics: Multivocal communication and religious appeals.
\newblock \emph{Political Behavior}, 37(1):3--26.

\bibitem[{Bhat and Klein(2020)}]{bhat2020covert}
Prashanth Bhat and Ofra Klein. 2020.
\newblock Covert hate speech: white nationalists and dog whistle communication
  on twitter.
\newblock In \emph{Twitter, the Public Sphere, and the Chaos of Online
  Deliberation}, pages 151--172. Springer.

\bibitem[{Bisk et~al.(2020)Bisk, Holtzman, Thomason, Andreas, Bengio, Chai,
  Lapata, Lazaridou, May, Nisnevich, Pinto, and Turian}]{bisk2020experience}
Yonatan Bisk, Ari Holtzman, Jesse Thomason, Jacob Andreas, Yoshua Bengio, Joyce
  Chai, Mirella Lapata, Angeliki Lazaridou, Jonathan May, Aleksandr Nisnevich,
  Nicolas Pinto, and Joseph~P. Turian. 2020.
\newblock Experience grounds language.
\newblock In \emph{{EMNLP}}.

\bibitem[{Chen et~al.(2019)Chen, Shen, Hu, Lu, Mei, and Liu}]{chen2019emoji}
Zhenpeng Chen, Sheng Shen, Ziniu Hu, Xuan Lu, Qiaozhu Mei, and Xuanzhe Liu.
  2019.
\newblock Emoji-powered representation learning for cross-lingual sentiment
  classification.
\newblock In \emph{{WWW}}.

\bibitem[{Clark et~al.(2018)Clark, Cowhey, Etzioni, Khot, Sabharwal, Schoenick,
  and Tafjord}]{arc}
Peter Clark, Isaac Cowhey, Oren Etzioni, Tushar Khot, Ashish Sabharwal, Carissa
  Schoenick, and Oyvind Tafjord. 2018.
\newblock Think you have solved question answering? try arc, the ai2 reasoning
  challenge.
\newblock \emph{arXiv preprint arXiv:1803.05457}.

\bibitem[{Dergousoff and Mandryk(2015)}]{game1}
Kristen~K. Dergousoff and Regan~L. Mandryk. 2015.
\newblock Mobile gamification for crowdsourcing data collection: Leveraging the
  freemium model.
\newblock In \emph{{CHI}}, pages 1065--1074. {ACM}.

\bibitem[{Devlin et~al.(2019)Devlin, Chang, Lee, and Toutanova}]{bert}
Jacob Devlin, Ming{-}Wei Chang, Kenton Lee, and Kristina Toutanova. 2019.
\newblock {BERT:} pre-training of deep bidirectional transformers for language
  understanding.
\newblock In \emph{{NAACL-HLT}}.

\bibitem[{Dieterich(1974)}]{dieterich1974public}
Daniel~J Dieterich. 1974.
\newblock Public doublespeak: Teaching about language in the marketplace.
\newblock \emph{College English}, 36(4):477--481.

\bibitem[{Grave et~al.(2018)Grave, Bojanowski, Gupta, Joulin, and
  Mikolov}]{grave2018learning}
Edouard Grave, Piotr Bojanowski, Prakhar Gupta, Armand Joulin, and Tomas
  Mikolov. 2018.
\newblock Learning word vectors for 157 languages.
\newblock In \emph{LREC}.

\bibitem[{Henderson and
  McCready(2019{\natexlab{a}})}]{henderson2019dogwhistles2}
Robert Henderson and Elin McCready. 2019{\natexlab{a}}.
\newblock Dogwhistles and the at-issue/non-at-issue distinction.
\newblock In \emph{Secondary content}, pages 222--245. Brill.

\bibitem[{Henderson and
  McCready(2019{\natexlab{b}})}]{henderson2019dogwhistles}
Robert Henderson and Elin McCready. 2019{\natexlab{b}}.
\newblock Dogwhistles, trust and ideology.
\newblock In \emph{Proceedings of the 22nd Amsterdam Colloquium}.

\bibitem[{Henderson and McCready(2020)}]{henderson-mccready-2020-towards}
Robert Henderson and Elin McCready. 2020.
\newblock Towards functional, agent-based models of dogwhistle communication.
\newblock In \emph{PaM}.

\bibitem[{Henderson and McCready(2017)}]{howdogwhistleswork}
Robert Henderson and Eric McCready. 2017.
\newblock How dogwhistles work.
\newblock In \emph{JSAI-isAI Workshops}, volume 10838 of \emph{Lecture Notes in
  Computer Science}, pages 231--240. Springer.

\bibitem[{Huang et~al.(2019)Huang, Ma, Lin, Han, and Hu}]{huang2019multimodel}
Longtao Huang, Ting Ma, Junyu Lin, Jizhong Han, and Songlin Hu. 2019.
\newblock A multimodal text matching model for obfuscated language
  identification in adversarial communication?
\newblock In \emph{{WWW}}.

\bibitem[{Huang et~al.(2017)Huang, Zhao, Lv, Lu, Zhai, and
  Hu}]{huang2017knowledge}
Longtao Huang, Lin Zhao, Shangwen Lv, Fangzhou Lu, Yue Zhai, and Songlin Hu.
  2017.
\newblock {KIEM:} {A} knowledge graph based method to identify entity morphs.
\newblock In \emph{{CIKM}}.

\bibitem[{Jin and Wu(2012)}]{semeval124}
Peng Jin and Yunfang Wu. 2012.
\newblock Semeval-2012 task 4: Evaluating chinese word similarity.
\newblock In \emph{SemEval@NAACL-HLT}.

\bibitem[{Kingma and Ba(2015)}]{adam}
Diederik~P. Kingma and Jimmy Ba. 2015.
\newblock Adam: {A} method for stochastic optimization.
\newblock In \emph{{ICLR}}.

\bibitem[{Lai et~al.(2017)Lai, Xie, Liu, Yang, and Hovy}]{race}
Guokun Lai, Qizhe Xie, Hanxiao Liu, Yiming Yang, and Eduard~H. Hovy. 2017.
\newblock {RACE:} large-scale reading comprehension dataset from examinations.
\newblock In \emph{{EMNLP}}.

\bibitem[{Lan et~al.(2020)Lan, Chen, Goodman, Gimpel, Sharma, and
  Soricut}]{albert}
Zhenzhong Lan, Mingda Chen, Sebastian Goodman, Kevin Gimpel, Piyush Sharma, and
  Radu Soricut. 2020.
\newblock {ALBERT:} {A} lite {BERT} for self-supervised learning of language
  representations.
\newblock In \emph{{ICLR}}.

\bibitem[{Li et~al.(2018)Li, Zhao, Hu, Li, Liu, and Du}]{analogical}
Shen Li, Zhe Zhao, Renfen Hu, Wensi Li, Tao Liu, and Xiaoyong Du. 2018.
\newblock Analogical reasoning on chinese morphological and semantic relations.
\newblock In \emph{ACL}.

\bibitem[{Liu et~al.(2018)Liu, Chen, Deng, Zeng, Chen, Li, and Tang}]{lcqmc}
Xin Liu, Qingcai Chen, Chong Deng, Huajun Zeng, Jing Chen, Dongfang Li, and
  Buzhou Tang. 2018.
\newblock {LCQMC:} {A} large-scale chinese question matching corpus.
\newblock In \emph{{COLING}}.

\bibitem[{Liu et~al.(2019)Liu, Ott, Goyal, Du, Joshi, Chen, Levy, Lewis,
  Zettlemoyer, and Stoyanov}]{roberta}
Yinhan Liu, Myle Ott, Naman Goyal, Jingfei Du, Mandar Joshi, Danqi Chen, Omer
  Levy, Mike Lewis, Luke Zettlemoyer, and Veselin Stoyanov. 2019.
\newblock Roberta: {A} robustly optimized {BERT} pretraining approach.
\newblock \emph{arXiv preprint arXiv:1907.11692}.

\bibitem[{L{\'o}pez(2015)}]{lopez2015dog}
Ian~Haney L{\'o}pez. 2015.
\newblock \emph{Dog whistle politics: How coded racial appeals have reinvented
  racism and wrecked the middle class}.
\newblock Oxford University Press.

\bibitem[{McArthur et~al.(2018)McArthur, Lam-McArthur, and
  Fontaine}]{mcarthur2018oxford}
Tom McArthur, Jacqueline Lam-McArthur, and Lise Fontaine. 2018.
\newblock \emph{Oxford companion to the English language}.
\newblock Oxford University Press.

\bibitem[{Pei(1973)}]{pei1973double}
Mario Pei. 1973.
\newblock Double-speak in america.

\bibitem[{Poliak et~al.(2018)Poliak, Naradowsky, Haldar, Rudinger, and
  Durme}]{sem18adam}
Adam Poliak, Jason Naradowsky, Aparajita Haldar, Rachel Rudinger, and
  Benjamin~Van Durme. 2018.
\newblock Hypothesis only baselines in natural language inference.
\newblock In \emph{*SEM@NAACL-HLT}.

\bibitem[{Prasetyo(2019)}]{prasetyo2019euphemism}
Ajeng~Ramadhani Prasetyo. 2019.
\newblock \emph{Euphemism Used by Trevor Noah in Stand-up Comedy Show Trevor
  Noah: Son of Patricia (2018)}.
\newblock Ph.D. thesis, Universitas Airlangga.

\bibitem[{Pruksachatkun et~al.(2020)Pruksachatkun, Phang, Liu, Htut, Zhang,
  Pang, Vania, Kann, and Bowman}]{intermediate}
Yada Pruksachatkun, Jason Phang, Haokun Liu, Phu~Mon Htut, Xiaoyi Zhang,
  Richard~Yuanzhe Pang, Clara Vania, Katharina Kann, and Samuel~R. Bowman.
  2020.
\newblock Intermediate-task transfer learning with pretrained language models:
  When and why does it work?
\newblock In \emph{{ACL}}.

\bibitem[{Pulley(1994)}]{pulley1994doublespeak}
Jerry~L Pulley. 1994.
\newblock Doublespeak and euphemisms in education.
\newblock \emph{The Clearing House}, 67(5):271--273.

\bibitem[{Sommerstein(1999)}]{sommerstein1999anatomy}
Alan~H Sommerstein. 1999.
\newblock The anatomy of euphemism in aristophanic comedy.
\newblock \emph{Studi sull’Eufemismo, Bari: Levante}, pages 183--217.

\bibitem[{Song et~al.(2018)Song, Shi, Li, and Zhang}]{tencentemb}
Yan Song, Shuming Shi, Jing Li, and Haisong Zhang. 2018.
\newblock Directional skip-gram: Explicitly distinguishing left and right
  context for word embeddings.
\newblock In \emph{{NAACL-HLT}}. Association for Computational Linguistics.

\bibitem[{Squires(2010)}]{squires2010enregistering}
Lauren Squires. 2010.
\newblock Enregistering internet language.
\newblock \emph{Language in Society}, 39(4):457--492.

\bibitem[{Sun et~al.(2019{\natexlab{a}})Sun, Qiu, and Huang}]{vcwe}
Chi Sun, Xipeng Qiu, and Xuanjing Huang. 2019{\natexlab{a}}.
\newblock {VCWE:} visual character-enhanced word embeddings.
\newblock In \emph{{NAACL-HLT}}.

\bibitem[{Sun et~al.(2019{\natexlab{b}})Sun, Wang, Li, Feng, Chen, Zhang, Tian,
  Zhu, Tian, and Wu}]{ernie}
Yu~Sun, Shuohuan Wang, Yukun Li, Shikun Feng, Xuyi Chen, Han Zhang, Xin Tian,
  Danxiang Zhu, Hao Tian, and Hua Wu. 2019{\natexlab{b}}.
\newblock Ernie: Enhanced representation through knowledge integration.
\newblock \emph{arXiv preprint arXiv:1904.09223}.

\bibitem[{Taylor(1974)}]{taylor1974terms}
Sharon~Henderson Taylor. 1974.
\newblock \href {http://www.jstor.org/stable/3087798} {Terms for low
  intelligence}.
\newblock \emph{American Speech}, 49(3/4):197--207.

\bibitem[{van Berkel et~al.(2017)van Berkel, Gon{\c{c}}alves, Hosio, and
  Kostakos}]{game2}
Niels van Berkel, Jorge Gon{\c{c}}alves, Simo Hosio, and Vassilis Kostakos.
  2017.
\newblock Gamification of mobile experience sampling improves data quality and
  quantity.
\newblock \emph{Proc. {ACM} Interact. Mob. Wearable Ubiquitous Technol.},
  1(3):107:1--107:21.

\bibitem[{Wolf et~al.(2020)Wolf, Debut, Sanh, Chaumond, Delangue, Moi, Cistac,
  Rault, Louf, Funtowicz, Davison, Shleifer, von Platen, Ma, Jernite, Plu, Xu,
  Scao, Gugger, Drame, Lhoest, and Rush}]{huggingface}
Thomas Wolf, Lysandre Debut, Victor Sanh, Julien Chaumond, Clement Delangue,
  Anthony Moi, Pierric Cistac, Tim Rault, R{\'{e}}mi Louf, Morgan Funtowicz,
  Joe Davison, Sam Shleifer, Patrick von Platen, Clara Ma, Yacine Jernite,
  Julien Plu, Canwen Xu, Teven~Le Scao, Sylvain Gugger, Mariama Drame, Quentin
  Lhoest, and Alexander~M. Rush. 2020.
\newblock Transformers: State-of-the-art natural language processing.
\newblock In \emph{{EMNLP} (Demos)}.

\bibitem[{Xu et~al.(2020{\natexlab{a}})Xu, Ge, Li, and Wei}]{unihanlm}
Canwen Xu, Tao Ge, Chenliang Li, and Furu Wei. 2020{\natexlab{a}}.
\newblock Unihanlm: Coarse-to-fine chinese-japanese language model pretraining
  with the unihan database.
\newblock In \emph{{AACL-IJCNLP}}.

\bibitem[{Xu et~al.(2020{\natexlab{b}})Xu, Pei, Wu, Liu, and Li}]{matinf}
Canwen Xu, Jiaxin Pei, Hongtao Wu, Yiyu Liu, and Chenliang Li.
  2020{\natexlab{b}}.
\newblock {MATINF:} {A} jointly labeled large-scale dataset for classification,
  question answering and summarization.
\newblock In \emph{{ACL}}.

\bibitem[{Xu et~al.(2020{\natexlab{c}})Xu, Zhou, Ge, Wei, and Zhou}]{bot}
Canwen Xu, Wangchunshu Zhou, Tao Ge, Furu Wei, and Ming Zhou.
  2020{\natexlab{c}}.
\newblock Bert-of-theseus: Compressing {BERT} by progressive module replacing.
\newblock In \emph{{EMNLP}}.

\bibitem[{Xu et~al.(2020{\natexlab{d}})Xu, Hu, Zhang, Li, Cao, Li, Xu, Sun, Yu,
  Yu, Tian, Dong, Liu, Shi, Cui, Li, Zeng, Wang, Xie, Li, Patterson, Tian,
  Zhang, Zhou, Liu, Zhao, Zhao, Yue, Zhang, Yang, Richardson, and Lan}]{clue}
Liang Xu, Hai Hu, Xuanwei Zhang, Lu~Li, Chenjie Cao, Yudong Li, Yechen Xu, Kai
  Sun, Dian Yu, Cong Yu, Yin Tian, Qianqian Dong, Weitang Liu, Bo~Shi, Yiming
  Cui, Junyi Li, Jun Zeng, Rongzhao Wang, Weijian Xie, Yanting Li, Yina
  Patterson, Zuoyu Tian, Yiwen Zhang, He~Zhou, Shaoweihua Liu, Zhe Zhao, Qipeng
  Zhao, Cong Yue, Xinrui Zhang, Zhengliang Yang, Kyle Richardson, and Zhenzhong
  Lan. 2020{\natexlab{d}}.
\newblock Clue: A chinese language understanding evaluation benchmark.
\newblock In \emph{{COLING}}.

\bibitem[{Zellers et~al.(2018)Zellers, Bisk, Schwartz, and Choi}]{swag}
Rowan Zellers, Yonatan Bisk, Roy Schwartz, and Yejin Choi. 2018.
\newblock {SWAG:} {A} large-scale adversarial dataset for grounded commonsense
  inference.
\newblock In \emph{{EMNLP}}.

\bibitem[{Zhang et~al.(2014)Zhang, Huang, Pan, Ji, Knight, Wen, Sun, Han, and
  Yener}]{zhang2014be}
Boliang Zhang, Hongzhao Huang, Xiaoman Pan, Heng Ji, Kevin Knight, Zhen Wen,
  Yizhou Sun, Jiawei Han, and B{\"{u}}lent Yener. 2014.
\newblock Be appropriate and funny: Automatic entity morph encoding.
\newblock In \emph{{ACL}}.

\bibitem[{Zhang et~al.(2015)Zhang, Huang, Pan, Li, Lin, Ji, Knight, Wen, Sun,
  Han, and Yener}]{zhang2015context}
Boliang Zhang, Hongzhao Huang, Xiaoman Pan, Sujian Li, Chin{-}Yew Lin, Heng Ji,
  Kevin Knight, Zhen Wen, Yizhou Sun, Jiawei Han, and B{\"{u}}lent Yener. 2015.
\newblock Context-aware entity morph decoding.
\newblock In \emph{{ACL}}.

\bibitem[{Zhang et~al.(2019)Zhang, Bai, Liang, Bai, Chang, Yu, Zhu, and
  Zhao}]{acl2019selection}
Guanhua Zhang, Bing Bai, Jian Liang, Kun Bai, Shiyu Chang, Mo~Yu, Conghui Zhu,
  and Tiejun Zhao. 2019.
\newblock Selection bias explorations and debias methods for natural language
  sentence matching datasets.
\newblock In \emph{{ACL}}.

\bibitem[{Zhou et~al.(2020)Zhou, Xu, Ge, McAuley, Xu, and Wei}]{pabee}
Wangchunshu Zhou, Canwen Xu, Tao Ge, Julian~J. McAuley, Ke~Xu, and Furu Wei.
  2020.
\newblock {BERT} loses patience: Fast and robust inference with early exit.
\newblock In \emph{NeurIPS}.

\end{thebibliography}
\bibliographystyle{acl_natbib}

\end{document}